\documentclass[journal]{IEEEtran}
\usepackage[utf8]{inputenc}
\usepackage{graphicx}

\usepackage{amssymb}
\usepackage{bm}
\usepackage{amsmath}
\usepackage{amssymb}
\usepackage[dvipsnames]{xcolor}
\definecolor{myblue}{RGB}{66, 114, 190}
\definecolor{myorange}{RGB}{240, 114, 38}
\usepackage{algpseudocode}
\usepackage{algorithm}  
\usepackage{algorithmicx} 

\usepackage{enumitem}

\usepackage{booktabs,multirow,multicol}
\usepackage[numbers,sort&compress]{natbib}
\usepackage[pagebackref=true,breaklinks=true,colorlinks,bookmarks=false]{hyperref}

\newcommand{\onedot}{.\null}
\newcommand{\etal}{\emph{et al}\onedot}
\def\eg{\emph{e.g}\onedot}

\newcommand{\revision}[1]{\textcolor{black}{#1}}

\begin{document}
%
\title{Mutual Context Network for Jointly Estimating Egocentric Gaze and Action}
%
%
%

\author{Yifei~Huang,
        Minjie~Cai,
        Zhenqiang~Li,
        Feng~Lu,~\IEEEmembership{Member,~IEEE,}
        and~Yoichi~Sato,~\IEEEmembership{Senior Member,~IEEE}
\thanks{Manuscript received * *, 2019; revised * *, *; accepted * *, *. This work was supported in part by JST CREST of Japan, and in part by National Natural Science Foundation of China (NSFC) under Grant 61906064 and 61972012. \textit{(Corresponding author: Minjie Cai)}}

\thanks{ Minjie Cai is with the College of Computer Science and Electronic Engineering, Hunan University, Changsha 410082, China (e-mail: caiminjie@hnu.edu.cn)}
\thanks{Yifei Huang, Zhenqiang Li and Yoichi Sato are with the Institute of Industrial Science, The University of Tokyo, Tokyo 1538505, Japan (e-mail: \{hyf,lzq,ysato\}@iis.u-tokyo.ac.jp)}
\thanks{Feng Lu is with the State Key Laboratory of VR system and technology, SCSE, Beihang University, Beijing 100191, China (e-mail: lufeng@buaa.edu.cn)}
}

\markboth{IEEE TRANSACTIONS ON IMAGE PROCESSING,~Vol.~*, No.~*, * *~2020}%
{Shell \MakeLowercase{\textit{et al.}}: Bare Demo of IEEEtran.cls for Journals}
%



\maketitle

\begin{abstract}
In this work, we address two coupled tasks of gaze prediction and action recognition in egocentric videos by exploring their mutual context\revision{: the information from gaze prediction facilitates action recognition and vice versa}. Our assumption is that during the procedure of performing a manipulation task, on the one hand, what a person is doing determines where the person is looking at. On the other hand, the gaze location reveals gaze regions which contain important and information about the undergoing action and also the non-gaze regions that include complimentary clues for differentiating some fine-grained actions. We propose a novel mutual context network (MCN) that jointly learns action-dependent gaze prediction and gaze-guided action recognition in an end-to-end manner. Experiments on multiple public egocentric video datasets demonstrate that our MCN achieves state-of-the-art performance of both gaze prediction and action recognition. Our experiments also show that action-dependent gaze patterns could be learned with our method.
\end{abstract}

\begin{IEEEkeywords}
gaze prediction, egocentric video, action recognition.
\end{IEEEkeywords}

%
\IEEEpeerreviewmaketitle

\section{Introduction}
\IEEEPARstart{T}{he} popularity of wearable cameras in recent years is accompanied by a large number of first-person-view videos, or often called egocentric videos, that record people's daily interactions with their surrounding environments. The demand for automatic analysis of egocentric videos has promoted various egocentric vision techniques \cite{betancourt2015evolution} such as egocentric video hyper-lapse \cite{kopf2014first, poleg2015egosampling} and video summarization \cite{lu2013story,xu2015gaze}. In particular, the task of understanding what a person is doing and where a person is looking at have attracted great interest from researchers. The former task is often called \textit{egocentric action recognition} \cite{li2015delving, ma2016going, yan2015egocentric} and the latter is called \textit{egocentric gaze prediction} \cite{li2013learning, zhang2017deep, huang2018predicting}. Although the two tasks have been studied extensively in the past years, few works have focused on the relationships between the two tasks which are in fact deeply related. 

In this work, we aim to jointly model the two coupled tasks of gaze prediction and action recognition in egocentric videos. \revision{Inspired by prior studies on visual attention during motor tasks~\cite{hayhoe2003visual,land2009vision,land2009looking} in psychology and cognitive science,} there are several previous works which studied how human gaze could benefit egocentric action recognition \cite{fathi2012learning, li2018eye}.
These work tried to model human gaze in egocentric videos and use estimated gaze points for removing unrelated background information. By paying more attention to the features extracted from the spatial regions approximate to the gaze location, improved action recognition performances could be achieved. This reveals the fact that the region of attention being looked at by a person provides discriminative information about what the person is doing. However, no previous effort of modeling has been seen to explore the inverse question: ``does what a person is doing affect the person's attention?"

\begin{figure}
    \centering
    \includegraphics[width=\linewidth]{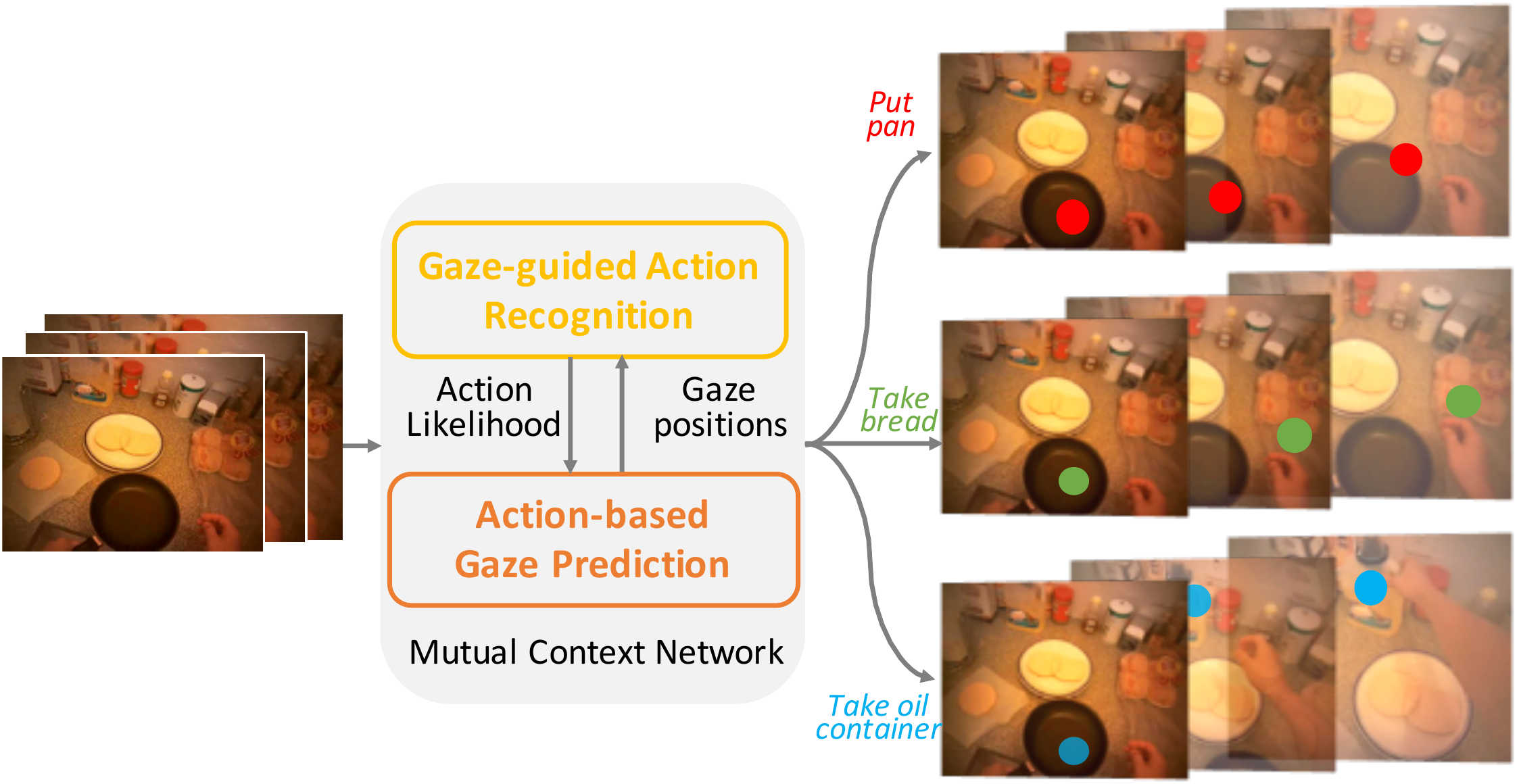}
    \caption{Illustration of our proposed mutual context network which takes egocentric video frames as input and jointly estimates action classes and gaze positions. The network models mutual context between egocentric gaze and action, with the motivation that information from one side facilitates the estimation of the other. For example, as shown in the right part of the figure, the predicted action class affects the estimation of gaze positions. Gaze positions are more likely to be estimated on the table if the action is predicted as ``put pan" (first row), while gaze positions would be more likely on the bread if the action is predicted as ``take bread" (second row).}
    \label{fig:concept}
\end{figure}

In an egocentric video, background regions are often cluttered and may contain multiple salient regions. Thus it would be difficult for a saliency-based model to predict gaze reliably without additional information about what a person is doing. \revision{Psychologists have investigated how current action being performed by a person implicitly affects gaze patterns on the same objects \cite{tipper1992selective,vickers2009advances,BelardinelliGoal}.} For example, to take a knife from a table, a person always moves his/her focus onto the knife and then keeps fixation on the knife before grasping it. Besides, different persons performing the same daily action (like ``put cup") often share similar gaze patterns. Therefore, we argue that for better modeling of gaze and actions in egocentric videos, not only the gaze-guided action recognition (gaze context for actions) but also the action-dependent gaze prediction (action context for gaze) should be jointly considered.


In this paper, we propose a deep learning-based framework that could jointly estimate human gaze positions and action classes in egocentric videos by modeling mutual context between the two coupled tasks. The framework is named as mutual context network (MCN). As illustrated in Figure \ref{fig:concept}, the proposed MCN takes video frames as input and outputs the likelihood of action classes as well as gaze positions for each frame. Two core modules for action-based gaze prediction and gaze-guided action recognition are newly developed within the MCN. The module of action-based gaze prediction leverages contextual information from the estimated action likelihood for gaze prediction, while the module of gaze-guided action recognition leverages contextual information from the predicted gaze positions for action recognition. The motivation of the two proposed modules is given in Section \ref{sec_motivation}.

Technically speaking, the action-based gaze prediction module takes the estimated action likelihood as input and produces a set of convolutional kernels which encode the semantic information relevant to the action being performed. This module also takes the feature maps encoded by a backbone network as input, and the generated kernels are then used to convolve with the input feature maps for locating action-related regions. We also take visual saliency into account and use a late fusion module to generate final gaze prediction which considers information from both low-level visual saliency and high-level action context. 
The gaze-guided action recognition module uses the predicted gaze positions as guidance to spatially aggregate the input features for action recognition. 
Specifically, the input feature maps are aggregated separately from the locations close to gaze position (gaze region) and the peripheral region (non-gaze region) and then combined in a selective manner, while the relative importance of the two regions is learned automatically during training.
More detailed description of MCN is given in Section \ref{sec_mcn}.
We conduct experiments on two public datasets: GTEA Gaze+~\cite{li2013learning} and EGTEA~\cite{li2018eye}. Experiments demonstrate that our method achieves state-of-the-art performance on both the gaze prediction task and the action recognition task.

Our main contributions are summarized as follows:
\begin{itemize}
    \item We propose a deep learning-based framework for both egocentric gaze prediction and action recognition that leverages the mutual context between the two tasks.
    \item We develop a novel action-based gaze prediction module which explicitly utilizes information from the estimated action likelihood for gaze prediction. To the best of our knowledge, this is the first work that considers action context for egocentric gaze prediction.
    \item Our proposed method achieves state-of-the-art performance on both gaze prediction and action recognition tasks.
\end{itemize}

\section{Related works}

\subsection{Egocentric gaze prediction}
Gaze prediction from egocentric video is a well-established research topic \cite{li2013learning} and can benefit a diverse range of applications such as action recognition \cite{fathi2012learning}, joint attention discovery \cite{huang2017temporal,kera2016discovering,park20123d,huang2020ego}, human computer interaction \cite{fujisaki2017interactive,khamis2016gazetouchpass,kurauchi2016eyeswipe}, and video summarization \cite{xu2015gaze}. 
Also, \revision{egocentric gaze cues can be used to infer cognitive states in developmental psychology studies} \cite{lawrence2017look,vernetti2018simulating,eckstein2017beyond}.
Despite the well-known correlation between gaze and saliency \cite{parkhurst2002modeling}, previous works have revealed the need for additional cues for predicting gaze in egocentric videos \cite{li2013learning,tavakoli2019digging,yamada2010can,yamada2011attention,zhang2017deep,zhang2018coarse}. Li \etal \cite{li2013learning} used multiple hand-crafted features such as head motion and hand configuration in a graphical model for gaze prediction in a cooking scenario. However, the pre-defined egocentric cues may limit the generalization ability of their model. Zhang \etal \cite{zhang2017deep} were the first to use deep learning for gaze prediction. Their method is similar to saliency prediction and tries to establish a mapping between image appearance and gaze positions. However, gaze prediction is challenging relying only on appearance, especially when the background is cluttered with multiple salient regions.
Huang \etal \cite{huang2018predicting} proposed a hybrid deep model that incorporates attention transition in addition to a bottom-up saliency-based model. \revision{However, they only modeled common patterns of attention transition for all actions and didn't consider the difference of gaze patterns in different actions.}

In this work, we explicitly leverage the contextual information from the performed actions for gaze prediction by using the predicted action likelihood. To the best of our knowledge, this is the first work to explore the influence of actions for egocentric gaze prediction.

\subsection{Egocentric action recognition}
Egocentric action recognition is one of the \revision{most focused research fields} in egocentric vision and has been studied extensively in recent years \cite{furnari2018leveraging,li2015delving,lu2019deep,mccandless2013object,ogaki2012coupling,poleg2016compact,pirsiavash2012detecting,spriggs2009temporal,surie2007activity,tang2017action,yonetani2016recognizing,cartas2018role,huang2020improving}. There are mainly two types of action recognition in egocentric vision: the first type is the coarse action recognition which aims to recognize the \emph{motion} of the camera wearer like ``put" or ``cut" \cite{kitani2011fast,singh2016first}. For example, Kitani \etal \cite{kitani2011fast} used global motion to discover different egocentric actions in an unsupervised manner. Singh \etal \cite{singh2016first} used deep learning with image, optical flow and additional inputs like hand masks to improve action recognition performance. Poleg \etal \cite{poleg2016compact} used 3D convolutions on optical flow images for long-term activity recognition. Another type of action recognition aims to recognize more fine-grained actions like ``take knife" and ``take cup", and attracts more research attention. For example, Fathi \etal \cite{fathi2011understanding} adopted a graphical model to recognize actions in relation to objects and head/hand motion. Ryoo \etal \cite{ryoo2015pooled} proposed a novel pooling method for action recognition. Ma \etal \cite{ma2016going} proposed a comprehensive deep model for recognizing objects and actions jointly. Sudhakaran \etal \cite{sudhakaran2018attention} used object-centric attention in a recurrent neural network to get better performance in action recognition. Wu \etal \cite{wu2019long} further used supportive information extracted over the entire span of a video called long-term feature bank to augment state-of-the-art action recognition models.
In this work, we focus on fine-grained action recognition. Different from previous work, our method recognizes actions with the contextual information from gaze by modeling actions and gaze in a unified framework.

\subsection{Gaze and actions}
Human gaze and actions are deeply correlated in egocentric videos, and the use of gaze has been proved to be beneficial for action recognition \cite{fathi2012learning,shen2018egocentric,zuo2018gaze}. For example, Li \etal \cite{li2015delving} used features extracted from gaze regions and improved the performance of action recognition. Shen \etal \cite{shen2018egocentric} encoded gaze with object proposal bounding boxes, defining gaze going inside and outside of a bounding box as an  ``event", and designed an asynchronous LSTM for action recognition. However, few works have been seen on the joint modeling of egocentric gaze prediction and action recognition. Extended from \cite{fathi2012learning}, Li \etal \cite{li2018eye} proposed a deep model for jointly modeling gaze and actions. They modeled the probabilistic nature of gaze and used the estimated gaze for better action recognition. However, their work did not explicitly consider the contextual information from actions for gaze prediction. We found that gaze prediction could be largely improved with the contextual information of actions. 

In this work, we leverage the mutual context of gaze and actions in our proposed model, in the form of using action likelihood as a conditional input to predict gaze and simultaneously, using gaze as guidance for action recognition. By explicitly exploring such mutual context, our model achieves the state-of-the-art performance in both gaze prediction and action recognition.

\section{Motivation}
\label{sec_motivation}

\begin{figure}
    \centering
    \includegraphics[width=\linewidth]{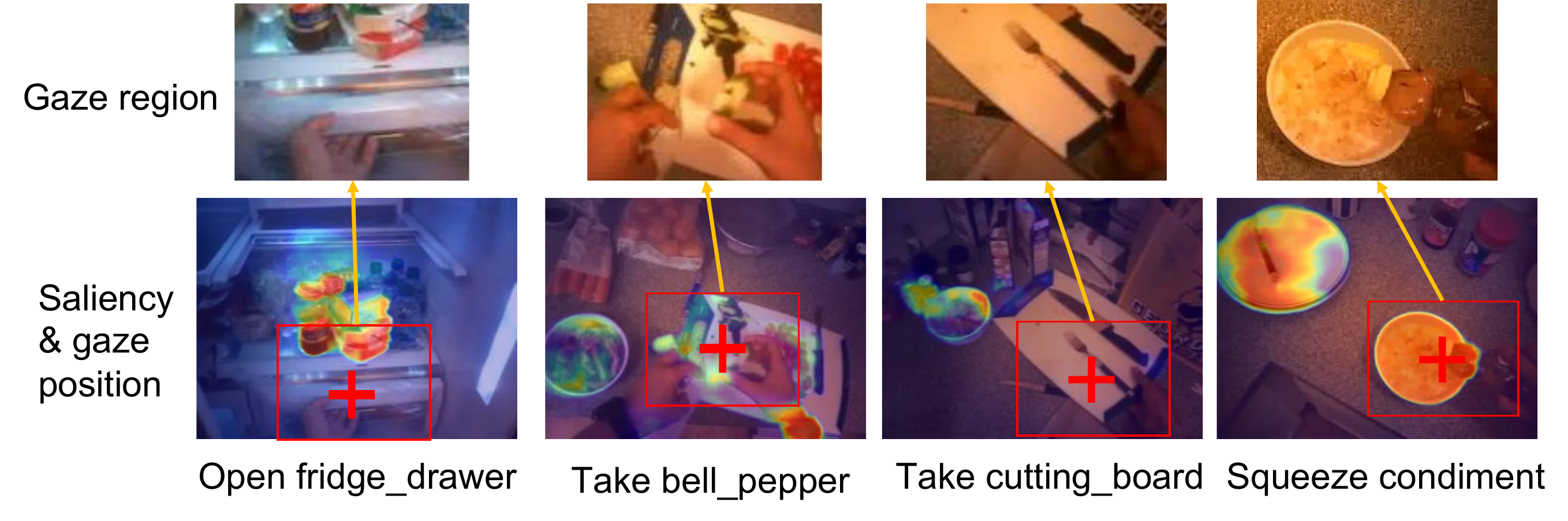}
    \caption{The difference between saliency maps (overlayed on images) and gaze regions (red cross, also enlarged above). The saliency maps are obtained using PiCANet \cite{liu2018picanet} pretrained on the DUTS dataset \cite{wang2017learning}. We can see that the gaze region is more action-dependent and can be significantly different from the visually salient regions.}
    \label{fig:motivation1}
\end{figure}

\subsection{Action context for gaze prediction}
\label{sec_motivation1}
To predict gaze positions from a video, it is important to locate the regions under human attention. Although visual saliency provides a way of extracting image regions that are likely to attract human attention, it may fail in egocentric videos which often contain cluttered background and thus multiple salient regions. To locate real gaze position from multiple ambiguous salient regions, we believe more discriminative information should be incorporated in the gaze prediction framework. \revision{Inspired by findings in psychology that different gaze patterns are produced for the same objects depending on different goals of actions~\cite{BelardinelliGoal},} we observed that there is a semantic connection between gaze region and the performed action, which could be used to improve the performance of gaze prediction. Egocentric action, which is composed of a verb and several nouns, encodes semantic information that is critical for locating the region of attention. For example, the nouns encode the region which is highly likely to be attended.
As shown in Figure \ref{fig:motivation1}, the gaze regions at the top row match well with the semantic information involved in the performed actions. For example, in the action of \textit{Open fridge\_drawer}, the object of \textit{fridge\_drawer} as well as the hand are contained in the real gaze region, while visual saliency is mainly distributed on other salient regions irrelevant to the performed action. 

Motivated by this semantic connection between the gaze region and the performed action, we propose a framework that could incorporate contextual information from action for gaze prediction. In the proposed framework, information about the performed action (\eg, represented by the softmax vector of action recognition) is used to produce intermediate information that is semantically meaningful and could be directly utilized for gaze prediction.

\subsection{Gaze context for action recognition}
\label{sec_motivation2}
\begin{figure}
    \centering
    \includegraphics[width=\linewidth]{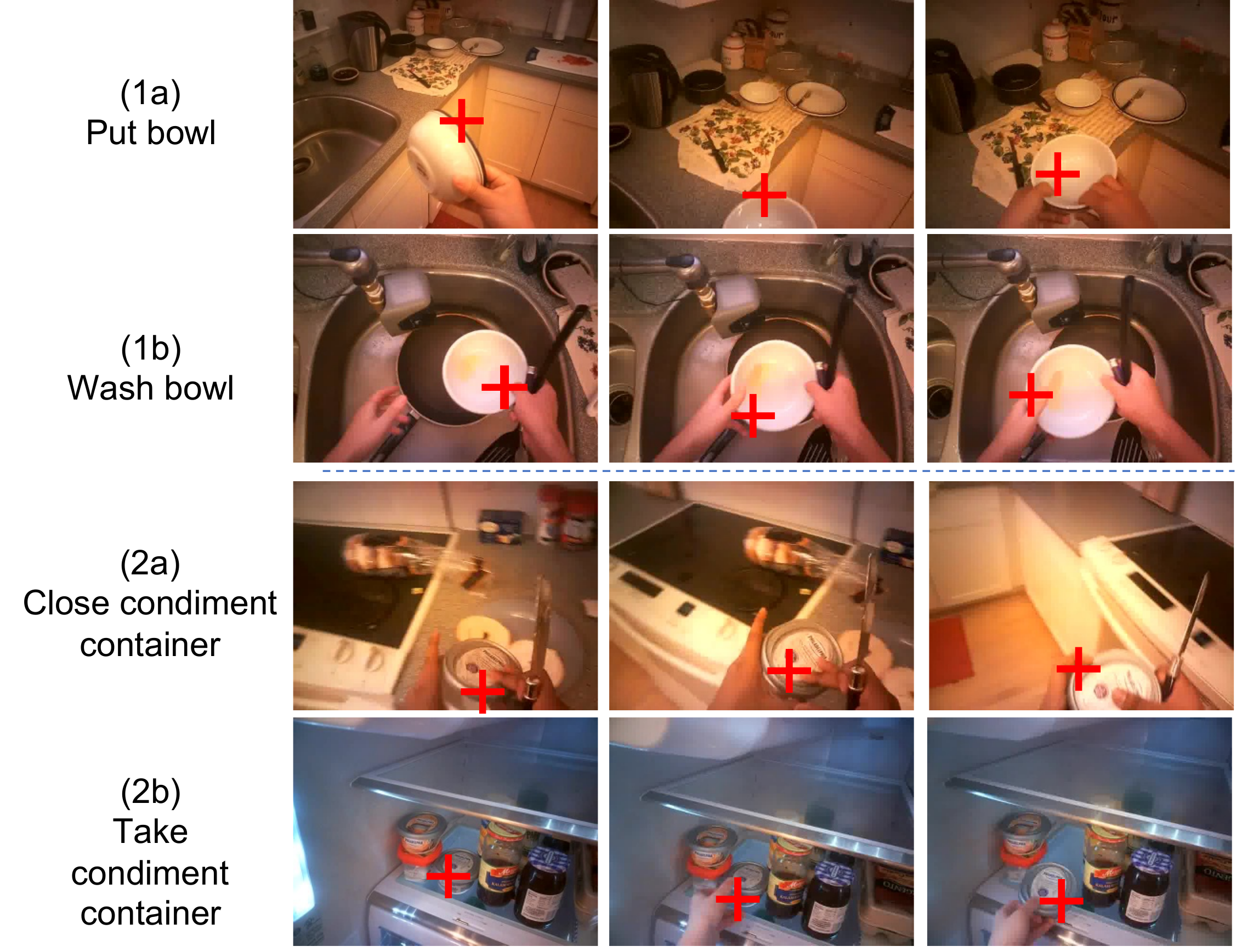}
    \caption{Gaze context for different actions. In (1a) and (1b), gaze focuses on the regions of bowl which help to recognize \textit{Put bowl} and \textit{Wash bowl} from other actions. With additional features from surrounding background, it is able to further differentiate the two actions. Similarly, in (2a) and (2b), it is easier to recognize \textit{Close condiment\_container} and \textit{Take condiment\_container} by extracting features from both gaze regions and background.}
    \label{fig:motivation2}
\end{figure}

It has been studied that humans use gaze to focus on important objects when performing an action \cite{hayhoe2005eye}, therefore, the region around human gaze reveals important information about the manipulated objects in egocentric actions. Based on such motivation, previous works \cite{fathi2011understanding,li2015delving,li2018eye} have shown that by focusing on visual features from gaze regions, it is possible to better recognize egocentric actions than using whole images. 
While information from the gaze regions is useful for recognizing many actions, the information from the surrounding background is also needed for differentiating some fine-grained actions with similar objects. As shown in (1a) and (1b) of Figure \ref{fig:motivation2}, it is hard to distinguish the actions only with information from gaze regions. The information from the surrounding background, in other words, the ``non-gaze" regions, helps to distinguish these two actions, since the existence of the sink in (1b) strongly indicates the action to be \textit{wash bowl} rather than \textit{put bowl}. Similarly in (2a) and (2b), the fridge and the containers around the gaze region help for the recognition of \textit{take condiment\_container} rather than \textit{close condiment\_container}. 

Thus, motivated by the usefulness of the regions guided by gaze in action recognition, we propose to make use of information from both the gaze regions and the non-gaze regions in a complementary way for better action recognition.

Overall, we propose a unified framework to model mutual context of action and gaze which could facilitate the two coupled tasks of egocentric gaze prediction and action recognition. The details are given in the following section.

\section{Mutual context network}
\label{sec_mcn}

\begin{figure*}
    \centering
    \includegraphics[width=\linewidth]{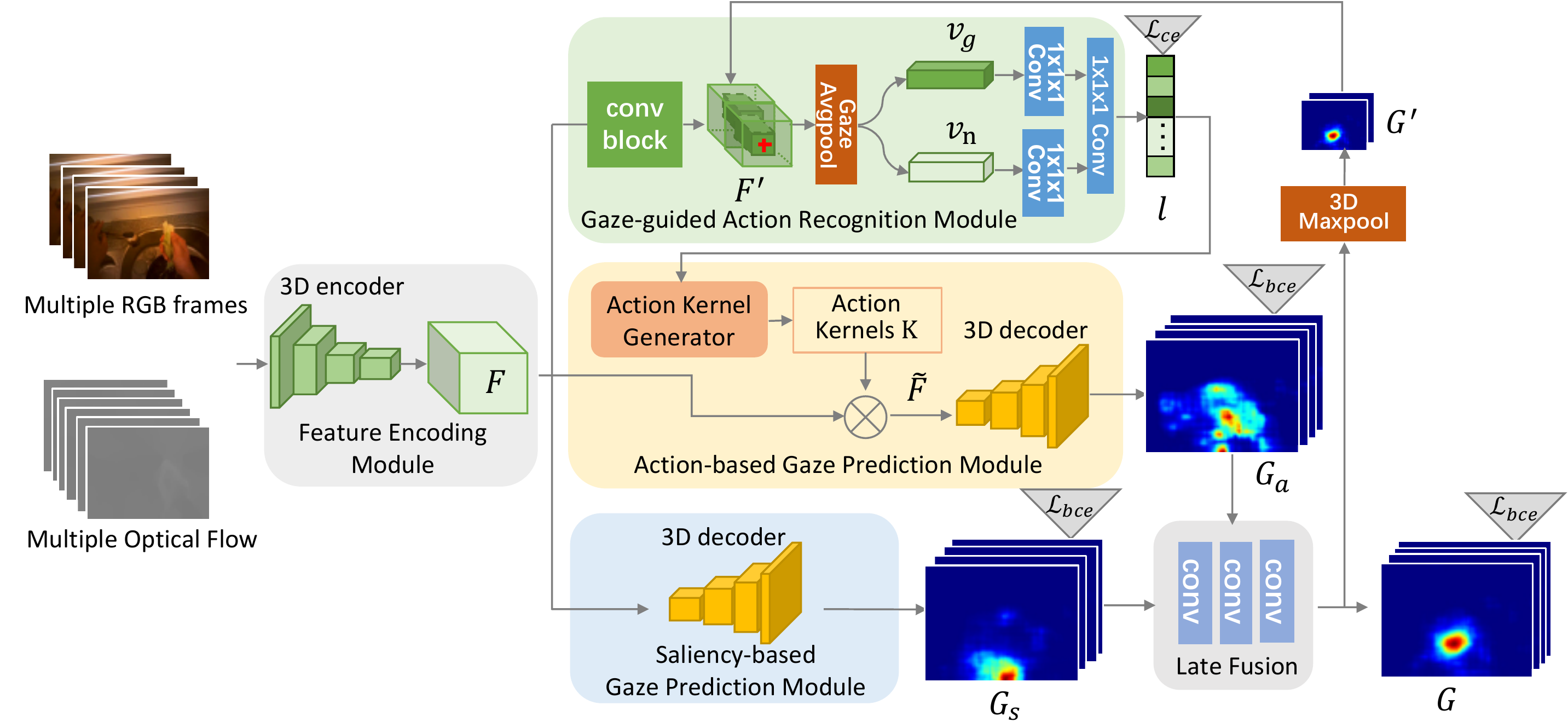}
    \caption{Architecture of our proposed mutual context network (MCN). MCN consists of 5 sub-modules: the feature encoding module which encodes input video frames into feature maps $F$, the gaze-guided action recognition module which uses gaze as a guideline to recognize actions, the action-based gaze prediction module which takes predicted action likelihood $l$ as input and outputs an action-dependent gaze probability map $G_a$, the saliency-based gaze prediction module which outputs a saliency map $G_s$, and finally the late fusion module to get the final gaze probability map $G$.}
    \label{fig:fullmodel}
\end{figure*}
 
\subsection{Overview}
In this work, we propose a mutual context network (MCN) that leverages the mutual context of action and gaze for joint gaze prediction and action recognition. The MCN uses the estimated action to predict the gaze point while in the meantime uses gaze as guidance for action recognition.

Figure \ref{fig:fullmodel} depicts the architecture of our MCN. 
The input video RGB frames and optical flow images are first encoded as feature maps $F$ by the feature encoding module, which are then used as input to the following modules.
One of the key components in our model is the action-based gaze prediction module that learns to predict gaze $G_a$ using the predicted action likelihood $l$ as a conditional input. As complementary information for gaze prediction, we also obtain a saliency map $G_s$ with the saliency-based gaze prediction module. The outputs from the two modules are then fused by the late fusion module to get the final gaze probability map $G = \{g_1, g_2, \cdots,g_N\}$.
Another component in our MCN is the gaze-guided action recognition module which takes the predicted gaze $G$ as guidance to selectively filter the input features for action recognition. The output of action likelihood $l$ is then used as conditional input to the action-based gaze prediction module, thus a loop of mutual context is constructed.

\subsection{Feature encoding module}
We adopt the first four convolutional blocks of the resnet50 version I3D network I3D-resnet \cite{wang2018non} for feature encoding. Following \cite{li2018eye}, we fuse the RGB stream and optical flow stream at the end of the 4th convolutional block by element-wise summation. With this 3D encoder, the output feature map $F$ is of size $(c, t, h, w)$, where $c$ is the number of channels, $t$ is the temporal dimension, and $(h,w)$ are the spatial height and width.

\subsection{Saliency-based gaze prediction module}
\revision{Saliency in image processing or computer vision community is often used to measure image regions that have unique and distinguishing appearance from the background such as a moving object or high contrast of brightness.} Image regions with high saliency tend to attract more human attention than other regions. Therefore, we use a saliency-based gaze prediction module to predict human attention from low-level image regions. For this, we use a 3D decoder that takes the encoded feature map $F$ as input and outputs a series of gaze probability maps $G_s$ with each pixel value within the range of [0, 1]. 
While this bottom-up approach provides information about salient regions in the image, it is not sufficient to reliably identify the attended region when multiple salient regions exist, which is common in egocentric videos.

\subsection{\revision{Gaze-guided action recognition module}}

Here we describe the gaze-guided action recognition module in our MCN that uses the predicted gaze point as a guide to exploit discriminative features for action recognition. Previous works \cite{fathi2012learning,li2018eye} mostly used gaze as a filter to remove features of image regions far from the gaze point. However, focusing only on the region around the gaze point might lose important information about the action. We observed that when performing certain actions such as ``put an object", the person may fixate on the table on which to place the object instead of looking at the object in hand which contains critical information about the action. Therefore, we think that while the gaze region is important, the region outside the gaze (non-gaze region) might also contain complementary information about the action. In this work, we develop a two-way pooling structure to aggregate features in the gaze and non-gaze regions separately and use both as input for action recognition.

As shown in Figure \ref{fig:fullmodel}, we first forward $F$ to the fifth convolutional block of I3D to encode more compact features $F' \in \mathbb{R}^{c'\times t'\times h'\times w'}$. On each temporal dimension of $F'$, we locate the corresponding spatial gaze point $(x_{\bar{t}}, y_{\bar{t}})$ on the feature map by selecting the spatial location of the maximum value in the 3d max-pooled gaze map $G'$. Then we split spatial dimensions of the feature map into two parts: gaze region and non-gaze region. Gaze region on a feature map (dark green region of $F'$ in the figure) is the locations whose spatial positions are within range $([x_{\bar{t}}-r,x_{\bar{t}}+r],[y_{\bar{t}}-r,y_{\bar{t}}+r])$, and non-gaze region is the left-out region (light green region of $F'$ in the figure). We pool the two regions separately on the spatial dimensions, generating two feature tensors $v_g$ and $v_n$:
\begin{gather}
    v_g[c,t] = \frac{\sum_{i=x_{\bar{t}}-r}^{x_{\bar{t}}+r}\sum_{j=y_{\bar{t}}-r}^{y_{\bar{t}}+r}\bar{F}'[c,t, i, j]}{4r^2} \\
    v_n[c,t] = \frac{\sum_i\sum_j \bar{F}'[c,t,i,j]-4r^2 v_g[c,t]}{h'\times w' - 4r^2},
\end{gather}
where $\bar{F}'_x[c,t,i,j]$ denotes the $c$-th channel and position $(t,i,j)$ of the feature map $F'_x$, similarly for $v[c,t]$.

The pooled feature tensors $v_g$ and $v_n$ are fed into two 1x1x1 convolution layers (denoted as $\mathcal{F}_g, \mathcal{F}_n$), and the outputs are channel-wise concatenated and forwarded into the final 1x1x1 convolution layer (denoted as $\mathcal{F}_{logit}$) for predictions. We average the predictions on temporal dimension to get the \revision{action likelihood $l\in \mathbb{R}^{n}$ of the $n$ action classes:}
\begin{gather}
v'_{g} \in \mathbb{R}^{s\times t} = \mathcal{F}_g(v_{g}) \\
v'_{n} \in \mathbb{R}^{s/2\times t} = \mathcal{F}_n(v_{n}) \\
l = Softmax(Average(\mathcal{F}_{logit}(\{ v'_{g} ; v'_{n} \})))
\end{gather}
Here $\{;\}$ denotes channel-wise concatenation. We set the output channel of $v'_{g}$ to be $s$ and $v'_{n}$ to be $\frac{s}{2}$ since the modeling of non-gaze region is empirically simpler than that of the gaze region, so we limit its channel size to prevent over-fitting.

\subsection{\revision{Action-based gaze prediction module}}

As different actions are associated with different objects and motion, people's gaze patterns when performing different actions are different. As stated in Section \ref{sec_motivation1}, motivated by the connection between the region of attention and the performed action, we propose an action-based gaze prediction module to leverage action information for more reliable gaze prediction. The proposed module is expected to be able to extract semantically meaningful information that could be used for locating gaze regions. To this end, inspired by \cite{visualdynamics18,chen2017stylebank}, we use the estimated action likelihood from the action recognition module to generate a group of convolutional kernels (called as ``action kernels'') which encode the semantic information of the performed action. 
The generated action kernels are then used to convolve the input features in order to locate the action-related regions. Finally, gaze probability maps that have the same size with input frames are generated by a decoder consisting of deconvolutional layers~\cite{badrinarayanan2015segnet,wu2018automatic,zhang2017deep}.

More formally, given action likelihood $l \in \mathbb{R}^{n}$ estimated by the action recognition module and the input feature maps $F \in \mathbb{R}^{c\times t\times h\times w}$ with $c$ channels ($t$ and $h,w$ are temporal and spatial dimension), the gaze probability map $G_a$ is generated through the following procedure: 
\begin{gather}
K = A(l)\\
\tilde{F} = K \otimes F \\
G_a = D(\tilde{F})
\end{gather}
where $A$ is the action kernel generator, $K \in \mathbb{R}^{k\times c\times k_t\times k_h\times k_w}$ is a group of $k$ kernels, and $\tilde{F} \in \mathbb{R}^{k\times t\times h\times w}$ is the filtered feature maps. $\otimes$ denotes the operator of convolution \revision{and $D$ denotes the decoder}. The kernel generator contains one fully connected layer and two convolutional layers. The output of the first fully connected layer is first reshaped into size $(k, k_t, k_h, k_w)$ and then forwarded to the following convolution layers.

We also adopt the saliency-based gaze prediction module which can be seen as a complementary to the action-based gaze prediction module. Finally, we use a late fusion module to combine the outputs $G_s$ and $G_a$ from the previous modules:
\begin{equation}
    G = L(G_s, G_a),
\end{equation}
\revision{where $L$ denotes the operation of the late fusion module.}
Late fusion technique has been proved to be effective in previous work of gaze prediction \cite{huang2018predicting}. Following previous works \cite{zhang2017deep,li2013learning}, we take the spatial location with maximum likelihood on $G$ as the predicted gaze point.

\subsection{Implementation and training details}
The whole framework is implemented using Pytorch framework \cite{paszke2017automatic}. The feature encoding module is identical to the first 4 convolutional blocks of the I3D-resnet \cite{wang2018non} network without the last pooling layer. The input of feature encoding module is 24 stacked images \revision{and corresponding dense optical flow images} with spatial size of $320 \times 240$, the output is feature maps with size of $c=1024, t=6, h=14, w =14$. \revision{Following Li \etal~\cite{li2018eye}, we use FlowNet~\cite{ilg2017flownet} to obtain dense optical flow images in $x/y$ form. The optical flow values are first truncated in the range of [-20, 20] and then rescaled to [0, 255].} The 3D decoder contains a set of 4 transposed convolution layers, with kernel sizes $4, 4, (3,4,4), (3,4,4)$, and stride $2,2,(1,2,2),(1,2,2)$ respectively. Padding 1 is added to all layers. Each layer is followed by batch normalization and ReLU activation. We add another convolution layer with kernel size 1 and a sigmoid layer on top of the 3D decoder for outputting values within $[0,1]$. The action kernel generator takes the input vector $l \in \mathbb{R}^{n}$ where $n$ is the number of action categories, and firstly encoded to a latent size of $\mathbb{R}^{4800}$ and reshaped into $(64,3,5,5)$. The two convolutional layers output channels $256$ and $1024$, with kernel size 3, stride 1 and padding 1. The output size of the action kernel generator is $(k,c, k_t,k_w,k_h)=(64,1024,3,5,5)$. For the gaze guided action recognition module, the convolution block is identical to the $5$-th convolution block of the I3D-resnet network. Thus the output size of $F_x'$ is $(c',t',h',w')=(1024,3,7,7)$. The 3d max-pooling layer therefore has kernel size (8,32,32). We set $r=1$ and $s=256$. The late fusion module is composed of 4 convolutional layers with output channels 32,32,8,1, in which the first 3 layers have a kernel size of 3 with 1 zero-padding and the last layer has a kernel size of 1 with no padding.

For training the whole network, we first train the gaze-guided action recognition module and the saliency-based gaze prediction module using ground truth (GT) action labels and gaze positions. We use Adam optimizer \cite{kingma2014adam} in all experiments.
The base I3D weights are initialized from weights pretrained on kinetics dataset \cite{kay2017kinetics}. We then use the result of action recognition to train the action-based gaze prediction module and then the late fusion module. We use cross entropy loss for action recognition and binary cross entropy loss for gaze prediction. We apply a Gaussian with $\sigma=18$ on the gaze point for generating ground truth images for gaze prediction. The learning rates for the action recognition module and all gaze prediction modules are fixed as $10^{-4}$ and $10^{-7}$ respectively. We first resize the images to $256\times 256$ and then random crop images into $224\times 224$, random flip with probability 0.5 for data augmentation during training. Ground truth gaze images perform the same data augmentation. 
\revision{Note that it is also possible to alternatively train the gaze-guided action recognition module and the action-based gaze prediction module using each other's output. However we found in the empirical study that training with GT information converges faster than with estimated information in an alternative manner. While alternative training could get slightly better action recognition performance and comparable gaze prediction performance, the training is much more time-consuming (nearly 6x time). Thus we did not adopt the alternative training strategy.} 

\begin{algorithm}[h!]  
  \caption{\small{Alternative inference procedure}}  
  \label{alg::training}  
  \begin{algorithmic}[1] 
  \small
    \State Using the saliency-based gaze prediction module to initialize gaze prediction $G$:
    \Statex ~$\qquad G \gets G_s$;
    \State Denote action likelihood vectors as $l$.
    \While {$e>0.1$ and $\#iteration\leq max\_iter$} 
      \State Update $l$ from gaze-guided action recognition module based on $G$;
      \State Get $G_a$ from action-based gaze prediction module using $l$;
      \State Update $G$ using $G_s$ and $G_a$: 
      \Statex ~$\qquad \quad G_{new} \gets LF(G_s, G_a)$;
      \State Compute the AAE of $G$ and $G_{new}$:
      \Statex ~$\qquad \quad e \gets AAE(G, G_{new})$;
      \State $G \gets G_{new}$
    \EndWhile
  \end{algorithmic}     
\end{algorithm}

When testing, we resize the image and send both the images and their flipped version and report the averaged performance. We iteratively infer gaze positions and action likelihood vectors in an alternative fashion as described in Algorithm \ref{alg::training}. The iteration terminates when the variation (measured by average angular error AAE) of current gaze prediction from the previous prediction is below a threshold or the number of iteration surpasses an upper bound. We empirically set this upper bound $max\_iter$ to be 10.

\section{Experiments}

\subsection{Dataset and evaluation metric}
Our experiments are conducted on two public datasets: EGTEA \cite{li2018eye} and GTEA Gaze+ \cite{li2013learning}. The GTEA Gaze+ dataset consists of 7 activities performed by 5 subjects. Each video clip is 10 to 15 minutes with resolution $1280\times960$. We do a 5-fold cross validation across all 5 subjects and take their average for evaluation as \cite{li2013learning}. The EGTEA dataset is an extension of GTEA Gaze+ which contains 29 hours of egocentric videos with the resolution of $1280\times960$ and 24 fps, taken from 86 unique sessions with 32 subjects performing meal preparation tasks in a kitchen environment. Fine-grained annotations of 106 action classes are provided together with measured ground truth gaze points on all frames. Following \cite{li2018eye}, we use the first split (8299 training and 2022 testing instances) of the dataset to evaluate the performance of gaze prediction and action recognition. We use the trimmed action clips of both datasets for training and testing unless otherwise noted.

We compare different methods on both tasks of gaze prediction and action recognition. For gaze prediction, we adopt two commonly used evaluation metrics: AAE (Average Angular Error in degrees) \cite{riche2013saliency} and AUC (Area Under Curve) \cite{borji2013analysis}. \revision{For action recognition, we use both instance level (inst.) and class level (cls.) classification accuracy~\cite{sudhakaran2019lsta} as the evaluation metric.}

\subsection{Gaze prediction results}
We compare our method with the following baselines:
\begin{itemize}
\item Saliency prediction methods: we use two representative traditional methods \textbf{GBVS} \cite{harel2007graph}, \textbf{Itti's model} \cite{itti2000saliency} as our baseline. We also re-implement the deep FCN based model \textbf{SALICON} \cite{huang2015salicon} as another baseline and train on the same dataset with gaze as ground truth saliency map.
\item Egocentric gaze prediction methods: We also compare with three egocentric gaze prediction methods closely to our work: coarse gaze prediction method (\textbf{Li \etal}\cite{li2018eye}), the GAN-based method (\textbf{DFG} \cite{zhang2017deep}), and the attention transition-based method (\textbf{Huang \etal} \cite{huang2018predicting}). Since \cite{li2018eye} only outputs a coarse gaze prediction map (of resolution $7\times7$), we resize their output using bilinear interpolation. For \textbf{Li \etal} and \textbf{DFG} we report the results based on our implementation as no code is publicly available. For \textbf{Huang \etal} we use the author's original implementation. 
\item Subsets of our full MCN: We also conduct ablation studies using subsets of our full model. These include the saliency-based gaze prediction module (\textbf{Saliency-based}), the action-based gaze prediction module (\textbf{Action-based}). In addition, we also test the action-based gaze prediction module with ground truth action labels (\textbf{Action-based$^{*}$}). To further validate that the action-based gaze prediction module can provide useful information, we change the action-based gaze prediction module to center-bias and feed them into the late fusion module, which forms the ablation baseline \textbf{Saliency-based + center bias}.
\end{itemize}

\begin{table}
\begin{center}
\begin{tabular}{lcccc}
\toprule
\multirow{2}{*}{Method} & \multicolumn{2}{c}{EGTEA} & \multicolumn{2}{c}{GTEA Gaze+}\\ \cmidrule{2-5}
 &AAE & AUC & AAE & AUC \\
\midrule
GBVS \cite{harel2007graph} & 12.81 & 0.707 & 12.68 & 0.829 \\
Itti \etal \cite{itti2000saliency}  & 12.50 & 0.717 & 12.73 & 0.801 \\
SALICON \cite{huang2015salicon} & 11.17 & 0.881 & 12.34 & 0.867\\
Li \etal \cite{li2018eye} & 8.58 & 0.870 & 8.97 & 0.889\\
DFG \cite{zhang2017deep} & 6.30 & 0.923 & 6.39 & 0.910\\
Huang \etal \cite{huang2018predicting} & 6.25 & 0.925 & 6.23 & 0.924\\
\midrule
Saliency-based & 6.36 & 0.922 & 6.57 & 0.929\\
Saliency-based + center bias & 6.30 & 0.924 & 6.51 & 0.930\\
Action-based  & 6.20 & 0.928 & 6.35 & 0.923\\
Action-based$^{*}$ & 6.04 & 0.927 & 6.20 & 0.933\\
Our full MCN & \textbf{5.79} & \textbf{0.932} & \textbf{5.74} & \textbf{0.945} \\
\bottomrule
\end{tabular}
\end{center}
\caption{Comparison of gaze prediction performance on two datasets. Results of previous methods are placed on top. Results of our full MCN and the subsets of MCN are placed on the bottom. Lower AAE and higher AUC indicate better performance. $^{*}$ denotes using ground truth action label as input.}
\label{aaetable}
\end{table}

Table \ref{aaetable} shows the quantitative comparison of different methods on gaze prediction performance. We first analyze the performance comparison with previous methods shown in the top part of the table. Our method outperforms state-of-the-art egocentric gaze prediction methods (\cite{zhang2017deep} and \cite{huang2018predicting}) on both datasets with the same experimental setting. It is important to notice that even our action-based gaze prediction module alone could achieve comparable performance with \cite{huang2018predicting}, which verifies the effectiveness of action context on gaze prediction.

\begin{figure*}
    \centering
    \includegraphics[width=\linewidth]{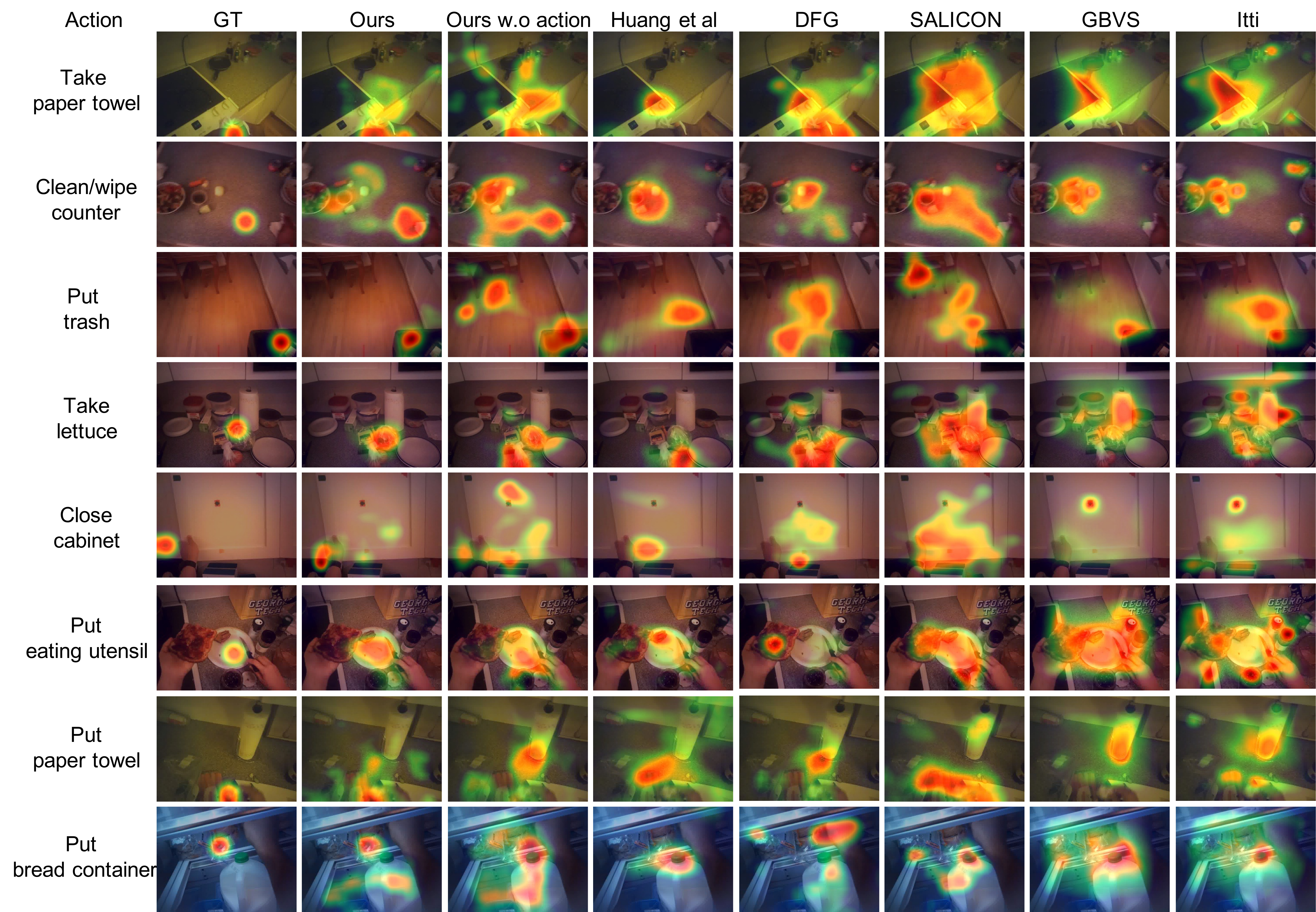}
    \caption{Qualitative visualizations of gaze prediction results on EGTEA dataset. We show the output heatmap from our full MCN and several baselines. Ground truth action labels and gaze points (\textbf{GT}) are placed on the leftmost columns.}
    \label{fig:gaze_quali}
\end{figure*}

We also conduct ablation studies by comparing different subsets of our MCN. As shown in the lower part of Table \ref{aaetable}, the action-based module performs better than the saliency-based module, verifying the effectiveness of action context in gaze prediction. When feeding the action-based module with ground-truth action labels, the performance is further improved. To examine the effectiveness of late fusion, we first tried the fusion of saliency-based module with center bias and found that it only slightly improves the performance of saliency-based module alone. However, the fusion of saliency-based module with action-based module (our full MCN) greatly improves the performance of two individual modules, as demonstrated by the decrease of AAE score from 6.36/6.20 to 5.79 on EGTEA dataset and from 6.57/6.35 to 5.74 on GTEA Gaze+ dataset. This indicates that an ideal gaze prediction method should consider information from both low-level visual saliency and high-level action context. 

Qualitative results are shown in Figure \ref{fig:gaze_quali}. It can be seen that with the help of the action-based gaze prediction module, our full MCN can better locate the action, thus giving better gaze prediction results. For example, in the first row, our MCN successfully recognizes the action as ``take paper towel'', thus finds the paper towel in the hand. Other baseline methods mostly focus on the stove or other salient regions. In the second row, while other methods are distracted by the plates and food on the counter, our MCN successfully locates the hand with dishrag on the bottom right corner and a part of the counter which will be cleaned in the next few frames. 
More interestingly as shown in the fourth row, the lettuce of ground-truth gaze fixation is placed on a cluttered kitchen table, which is challenging for other methods to locate. Still, our full MCN correctly predicts gaze to be on the lettuce with the help of context from the action ``take lettuce''.  Similar situations can be found in other rows of the figure.

\subsection{Examination of action-based gaze prediction module}
We further demonstrate that our action-based gaze prediction module is able to learn meaningful gaze patterns relevant to different actions. Intuitively, the gaze patterns for similar actions should also be similar: for example, for the action ``take bowl", the gaze prediction performance should not decrease obviously if we use a label of ``take plate" as input to the action-based gaze prediction module, but should decline sharply if it is given the label of ``cut tomato" as input.
Thus we conduct a new experiment on the top 20 frequent actions in the test set of EGTEA dataset to examine our action-based gaze prediction module. We feed the module with action labels representing each of the 20 action classes and examine how gaze prediction performance (AAE score) varies when the module is tested on each of these actions. For example, we feed the action-based gaze prediction module with the action label of ``take plate" and test the AAE scores on the videos of all 20 actions. As a result, we obtain a matrix of AAE scores with the size of $20\times20$, denoted by $M$, in which $M_{i,j}$ is the AAE score of the action-based gaze module fed with the action label of the $i$-th action and applied to the videos of the $j$-th action.

We found that the average AAE on the diagonal of $M$ is 6.21, while the average AAE of $M$ without diagonal is 6.87. This indicates that the correct action label can benefit the predictions of action-based gaze prediction module. To better understand the effect of different action labels on the action-based gaze prediction module, we visualize an ``affinity matrix" $A$ with the following equation:
\begin{equation}
    A_{i,j} = 1 - \frac{M_{i,j}-min(M_{i,*})}{max(M_{i,*})-min(M_{i,*})},
\end{equation}
where $A_{i,j}$ can be seen as the ``affinity score" between the measured ground truth gaze pattern of the $i$-th action and the learned gaze pattern of the $j$-th action. We normalize each number to have numeric range of [0,1]. 

\begin{figure}
    \centering
    \includegraphics[width=\linewidth]{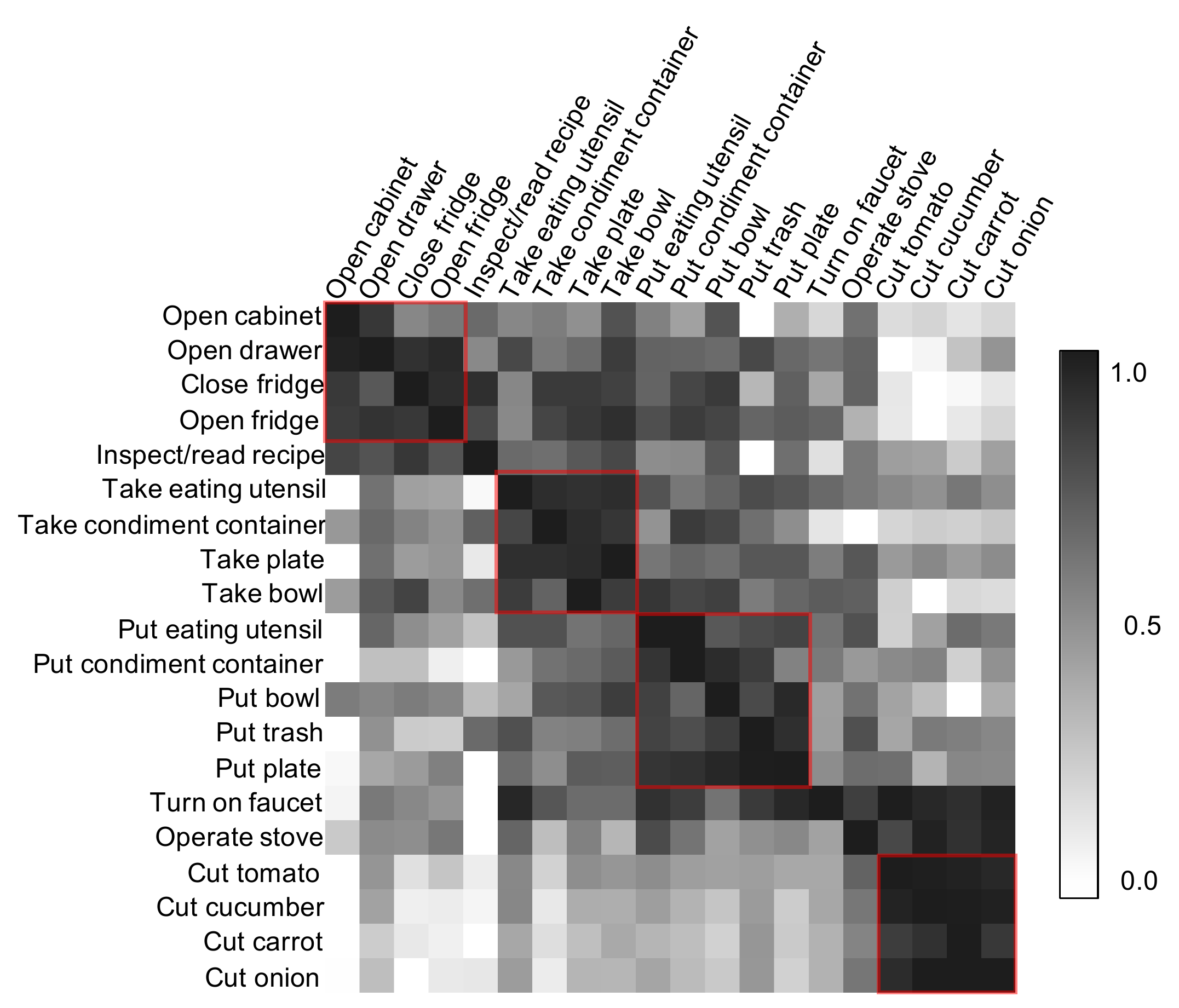}
    \caption{Affinity matrix of the top 20 frequent actions in EGTEA dataset. Actions are re-ordered for the ease of viewing. Each row of the matrix represents the ``affinity score" of one action against all the 20 actions. Darker indicates higher ``affinity" between corresponding actions. We mark several darker groups of similar action with high ``affinity" for the ease of reading.}
    \label{fig:confusion}
\end{figure}
We visualize the affinity matrix in Figure \ref{fig:confusion}. We can see from several dark blocks along the diagonal (marked by boxes) that there exist several groups of actions of which the learned gaze patterns are similar to each other, for example, the action group of ``put" in the middle and the action group of ``cut" on the bottom-right of the figure. The obtained affinity matrix is actually consistent with our common sense of these actions. For the action group of ``put'', persons tend to fixate on a table which is often the destination of placement. For the action group of ``cut'', the gaze is often fixated on a knife. More importantly, the results show that our action-based gaze prediction module has learned meaningful action-based gaze patterns. We think these patterns might be used to study the similarity between different actions from the perspective of human attention in future works.

\subsection{Action recognition results}

As for the task of action recognition, we compare our method with the following methods:
\begin{itemize}
\item \textbf{I3D} \cite{carreira2017quo} is one of the state of the art models for action recognition. We refer to \cite{li2018eye} for the accuracy of this baseline method.
\item Methods using measured gaze: \textbf{I3D+Gaze} is to use a ground truth gaze point as a guideline to pool feature maps from the last convolution layer of the fifth convolutional block. \textbf{EgoIDT+Gaze} \cite{li2015delving} is a traditional method which uses dense trajectories \cite{wang2011action} selected by a ground truth gaze point for action recognition.
\item State-of-the-art egocentric action recognition methods: \textbf{Li \etal} \cite{li2018eye} uses a estimated gaze probability map as soft attention to perform a weighted average on top I3D features. \textbf{Sudhakaran \etal} \cite{sudhakaran2018attention} adopts attention mechanism in a recurrent neural network to recognize actions. \textbf{LSTA} \cite{sudhakaran2019lsta} is a recent RNN based egocentric action recognition method that models action related attention for better action recognition. We also compare our method with \textbf{Ma \etal} \cite{ma2016going} and \textbf{Shen \etal} \cite{shen2018egocentric} that use additional annotations of object locations and hand masks during training. \cite{shen2018egocentric} even uses ground-truth gaze positions as input during testing. We compare the performance as reported in their original papers.
\item Baselines of our model: \textbf{MCN (w/o gaze)} is the baseline that does not use gaze information and is constructed to validate the effectiveness of the gaze-guided action recognition module. It performs a direct average pooling as in \cite{carreira2017quo, wang2018non}. The \textbf{MCN (center bias)} is the baseline that uses the image center as the predicted gaze position. We construct this baseline to validate the usefulness of better gaze prediction on action recognition. \textbf{MCN (gaze region)} is a baseline of our MCN that uses only the gaze-centered region for pooling. We use this baseline to validate the usefulness of information from the non-gaze regions. \textbf{MCN (soft gaze)} is a baseline that uses the predicted gaze probability map as a soft attention map on the features of the final convolutional block as in \cite{li2018eye}.
\end{itemize}

\begin{table}
\begin{center}
\begin{tabular}{lcccc}
\toprule
\multirow{2}{*}{Method} & \multicolumn{2}{c}{EGTEA} & \multicolumn{2}{c}{GTEA Gaze+} \\
 & inst. & cls. & inst. & cls. \\
\midrule
EgoIDT + Gaze \cite{li2015delving}& - & 46.5 & - & 60.5 \\
I3D \cite{carreira2017quo} &  - & 49.8 & - & 57.6 \\
I3D \cite{carreira2017quo} + Gaze &  - & 51.2 & - & 59.7 \\
Li \etal \cite{li2018eye} &  - & 53.3 & - & -\\
Ma \etal \cite{ma2016going} & -&- & 66.4 & - \\
Shen \etal \cite{shen2018egocentric} & - & - & 67.1 & - \\
Sudhakaran \etal \cite{sudhakaran2018attention} & 60.8 & 52.4 & 60.1 & 55.4 \\
LSTA \cite{sudhakaran2019lsta} & 61.9 & 53.0 & - & - \\
\midrule
MCN (w/o gaze) & 55.6 & 45.5 & 59.9 & 55.4 \\
MCN (center bias) & 53.2 & 43.3 & 59.6 & 55.1 \\
MCN (gaze region) & 56.4 & 45.9 & 61.1 & 56.0 \\
MCN (soft gaze) & 60.8 & 51.7 & 65.5 & 59.9 \\
Our full MCN & \textbf{62.6} & \textbf{55.7} & \textbf{67.4} & \textbf{61.5} \\
\bottomrule
\end{tabular}
\end{center}
\caption{Quantitative comparison of action recognition. \revision{We report instance level accuracy (inst.) and class level accuracy (cls.) in \%.} Values in brackets indicate the methods that rely on ground truth gaze.}
\label{acctable}
\end{table}

Quantitative comparison of different methods on two datasets is shown in Table \ref{acctable}. The deep learning method I3D \cite{carreira2017quo} outperforms EgoIDT+Gaze \cite{li2015delving} that uses handcrafted features on EGTEA dataset but not on GTEA Gaze+ dataset. This is possibly due to the smaller number of training samples in GTEA Gaze+ dataset. With the use of measured gaze, the performance of I3D+Gaze is improved compared with I3D. On both datasets, our MCN outperforms state-of-the-art methods (\cite{ma2016going}\cite{shen2018egocentric}\cite{sudhakaran2019lsta}), including \cite{shen2018egocentric} that relies on ground-truth gaze positions during testing. 

We also conduct an ablation study to examine the effectiveness of different components of our model. The baseline of (MCN w/o gaze) takes whole images as input without considering distinct information from gaze or non-gaze regions. It performs better than the similar method of I3D \cite{carreira2017quo} and shows the advantage of the more advanced base network (I3D-resnet \cite{wang2018non}) adopted in our model. The comparison between MCN (center bias) and (gaze region) indicates the usefulness of predicted gaze for action recognition. The superiority of our full model over MCN (gaze region) indicates the usefulness of the non-gaze regions, and validates our thought that the non-gaze regions should be considered together with gaze regions in action recognition. Although MCN (soft gaze) partly considers regions distant from gaze with less weight, our full model outperforms MCN (soft gaze) by explicitly incorporating information from gaze and non-gaze regions.

\begin{figure}
    \centering
    \includegraphics[width=\linewidth]{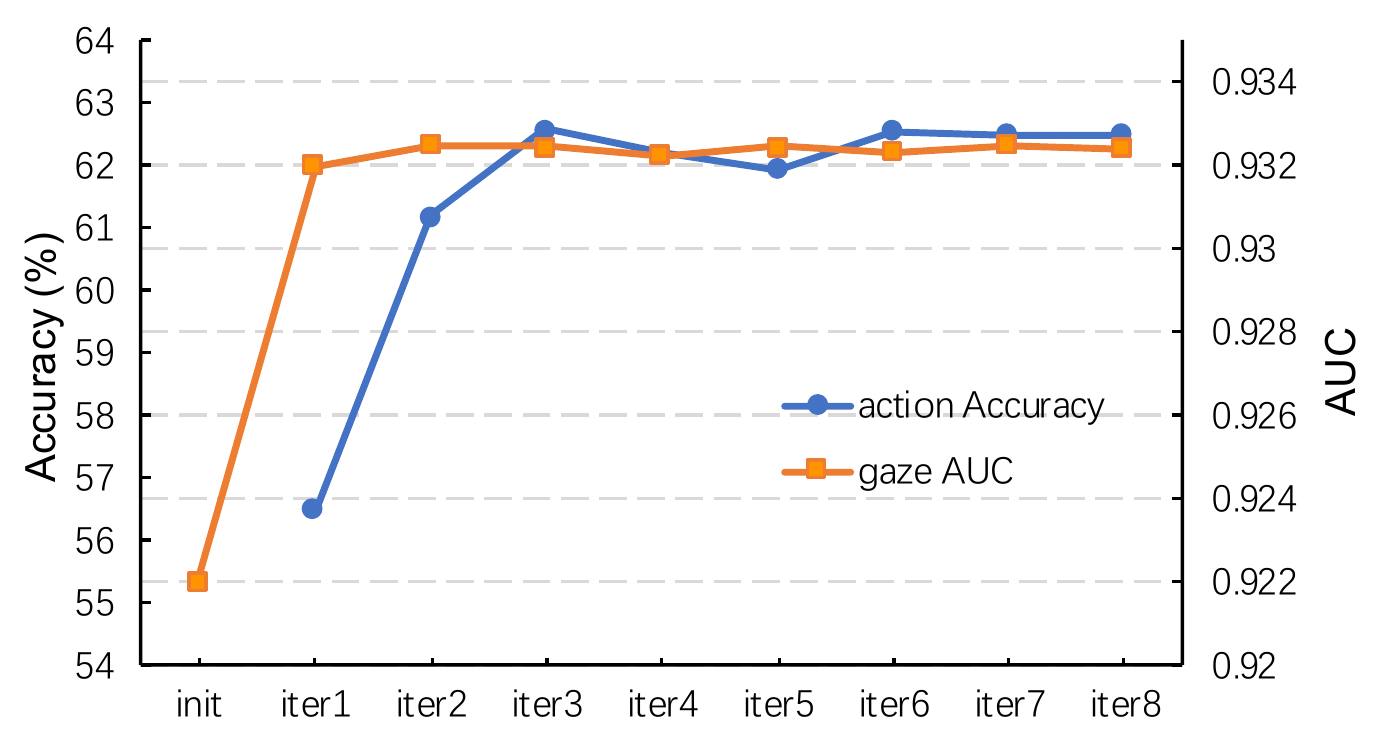}
    \caption{Gaze prediction AUC and action recognition accuracy with respect to inference iteration on the EGTEA dataset. Blue curve with circle markers correspond to action recognition accuracy on the left axis, and orange curve with square markers correspond to gaze prediction AUC on the right axis. }
    \label{fig:iteration}
\end{figure}

\section{Discussion}
\subsection{Model convergence}

In the proposed method, the network inference is conducted in an alternative manner. Here, we show the performance of \revision{action recognition and gaze prediction} on EGTEA dataset with our method at different iterations of inference in Figure \ref{fig:iteration}. Note that at the stage of initialization, only the saliency-based module is used. Then gaze-guided action recognition and action-based gaze prediction are conducted alternatively from the first iteration. It can be seen that the performance of both gaze prediction and action recognition increases dramatically at first and converges after about two iterations. This strongly supports our hypothesis that the mutual context of gaze and action can be beneficial for both tasks. In addition, the performance of gaze prediction converges faster than that of action recognition. We think the reason might be that even coarse information of actions (\eg, the object or verb of an action) is sufficient as the context for gaze prediction. Actually, this is also demonstrated as in Figure \ref{fig:confusion} that several groups of actions have learned similar gaze patterns.

\revision{Note that the computational cost for one round of iterative inference in our method is 128.2 Giga floating point operations (GFLOPs). With the same backbone, the computational cost for independent action recognition (83.5 GFLOPs) and gaze prediction (83.6 GFLOPs) takes in total 167.1 GFLOPs. Since our model almost converges with two iterations, the total computational cost for our method is 256.4 GFLOPs, which is around 1.5 times that of independent inference.}

\subsection{Failure cases}

\begin{figure}
    \centering
    \includegraphics[width=\linewidth]{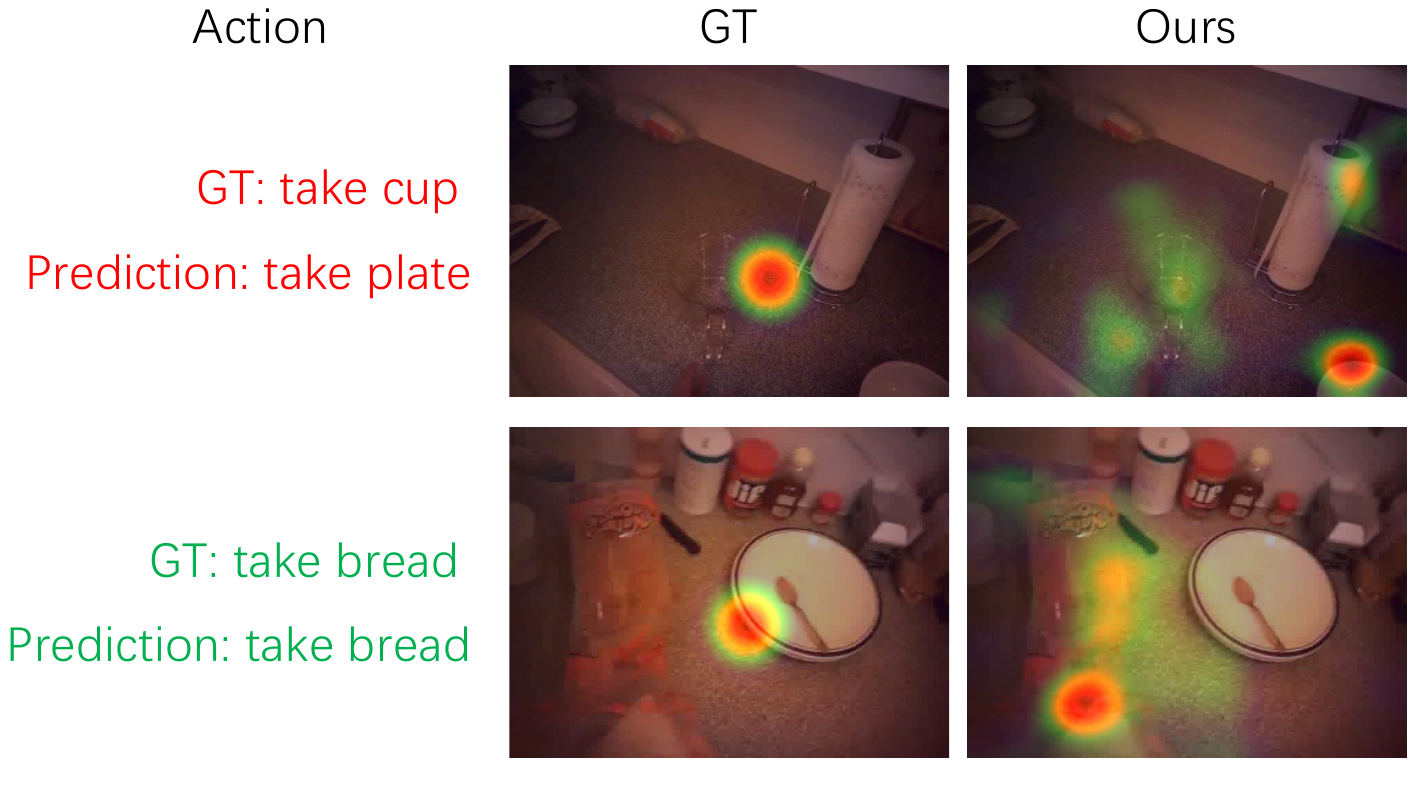}
    \caption{Failure cases of our MCN on gaze prediction. In the first row, failed action recognition misleads gaze prediction. In the second row, although the action recognition is correct, the camera wearer shifts the gaze fixation onto the region of future destination when he/she has already finished the action of grabbing the bread.}
    \label{fig:fail}
\end{figure}

Here we discuss several failure cases of gaze prediction with our method. One failure case happens when critical information of an action is incorrectly predicted. As shown in the first row of Figure \ref{fig:fail}, the wrong prediction of ``take cup'' as ``take plate'' causes our model to focus on the region of the plate while true gaze is on the region of the cup. Still, the impact of failed action recognition is limited in our model. We have analyzed gaze prediction results in two opposite cases and found that: among all the testing data of the EGTEA dataset, our model achieves an AAE score of $6.01$ when action recognition fails, and an AAE score of $5.68$ when action recognition is correct.

Another failure case comes from the circumstances when a person begins to shift the gaze fixation between consecutive actions. An example is shown in the second row of Figure \ref{fig:fail}. After grabbing the bread, instead of keeping fixation on the bread, the person's attention goes to the plate on which he's planning to put the bread. \revision{Actually such case has also been investigated on the topic of eye-hand coordination by psychologists~\cite{hayhoe2003visual,land2009vision,land2009looking}}.
Our current method is trained based on trimmed action sequences and could not identify such circumstances, thus fails to predict the true gaze positions at the boundaries of consecutive actions. This reveals the necessity of taking attention transition into consideration for our current gaze prediction model.

To further analyze the effect of learning attention transition with untrimmed video, we also compared the performance of the original version of \cite{huang2018predicting} (denoted as Huang \etal\dag) which is trained based on untrimmed videos on the GTEA Gaze+ dataset. By learning attention transition, Huang \etal\dag~ achieves the performance of 4.83 in the AAE metric and 0.939 in the AUC metric and outperforms our method by the metric of AAE (4.83 versus 5.74).
Meanwhile, when trained with the same trimmed videos, Huang \etal ~\cite{huang2018predicting} clearly performs worse than our method (as shown in Table \ref{aaetable}), possibly due to the lack of consideration for action context. 
Overall, the comparison between our method and the two variants of \cite{huang2018predicting} shows that while our method can benefit from action context and achieves state-of-the-art performance on the trimmed dataset, its current version could not capture temporal evolution of attention in each action. We think the combination of action-based gaze prediction and attention transition could be a good research direction to explore. \revision{Although Huang \etal\dag ~\cite{huang2018predicting} modeled attention transition between consecutive sub-goals of actions, they assumed common patterns of attention transition in different actions. As investigated by Belardinelli \etal~\cite{BelardinelliGoal} that gaze patterns on the same set of objects are different depending on the action being performed, therefore we believe attention transition should be modeled in the context of actions. We would further explore this research direction in our future work.}

\section{Conclusion and future work}
\label{sec_conclusion}
In this work, we proposed a novel deep model for both egocentric gaze prediction and action recognition. Our model explicitly leverages the mutual context between the two tasks. Within our model, the action-based gaze prediction module predicts gaze positions using a set of convolutional kernels generated based on the action likelihood. The gaze-guided action recognition module selectively aggregates the features of gaze region and non-gaze region for better action recognition. Experiments show that our model achieves state-of-the-art performance for both tasks on two public egocentric video datasets.

Although our model outperforms previous methods in trimmed action sequences, gaze prediction performance still needs further improvement, especially for the transition periods between consecutive actions in untrimmed videos. As for the future work, We think it would be an interesting direction to explore the gaze transition patterns in a broader activity scope which involves multiple consecutive fine-grained actions. Another possible direction of future work is to study the ways of mitigating the negative influence of failed action recognition on gaze prediction. We think considering the likelihood of the top-5 action and their relations might be a promising direction to explore. 

\section*{Acknowledgement}
This work was supported by JST CREST Grant Number JPMJCR14E1 and JST AIP Acceleration Research Grant Number JPMJCR20U1, Japan.

\ifCLASSOPTIONcaptionsoff
  \newpage
\fi



%
\bibliographystyle{IEEEtran}
\bibliography{bib}
%


\vfill


\end{document}